\documentclass[journal]{IEEEtran}

\IEEEoverridecommandlockouts       

\usepackage{graphics} 
\usepackage{color}
\usepackage{xcolor}
\usepackage{cite}
\makeatletter
\let\NAT@parse\undefined
\makeatother
\usepackage[colorlinks,citecolor=blue,urlcolor=blue,bookmarks=false,hypertexnames=true]{hyperref} 
\usepackage{float}
\usepackage{makecell}
\usepackage{longtable}
\usepackage{stfloats}
\usepackage{multirow}
\usepackage{amsmath}
\usepackage{graphicx}
\usepackage{subcaption}
\usepackage{bm}
\usepackage{threeparttable}
\usepackage{balance}
\usepackage[linesnumbered,ruled]{algorithm2e}
\usepackage[numbers,sort&compress]{natbib}

\title{
On the Federated Learning Framework for Cooperative Perception
}

\author{\IEEEauthorblockA{
Zhenrong Zhang\IEEEauthorrefmark{1},
Jianan Liu\IEEEauthorrefmark{1},
Xi Zhou,
Tao Huang,~\IEEEmembership{Senior Member,~IEEE,}\\
Qing-Long Han,~\IEEEmembership{Fellow,~IEEE}, 
Jingxin Liu,~\IEEEmembership{Member,~IEEE}, and 
Hongbin Liu,\IEEEauthorrefmark{2}~\IEEEmembership{Member,~IEEE}
}
\vspace{-5 mm}

\thanks{\IEEEauthorrefmark{1}Both authors contribute equally to the work and are co-first authors.}
\thanks{\IEEEauthorrefmark{2}Corresponding author.}
\thanks{The works of Z. Zhang, J.X.~Liu and H. Liu were jointly supported by the National Natural Science Foundation of China (62201474), Suzhou Science and Technology Development Planning Programme (Grant No.ZXL2023171) and XJTLU Research Development Fund (RDF-22-01-129, RDF-21-02-084).}
\thanks{Z.~Zhang, H.~Liu and J.X.~Liu are with the School of AI and Advanced Computing, Xi'an Jiaotong-Liverpool University, Suzhou, P.R.~China. Email: zhenrong.zhang21@student.xjtlu.edu.cn, \{hongbin.liu, jingxin.liu\}@xjtlu.edu.cn.}
\thanks{J.~Liu is with Momoni AI, Gothenburg, Sweden. Email: jianan.liu@momoniai.com.}
\thanks{X.~Zhou and T.~Huang are with the College of Science and Engineering, James Cook University, Cairns, QLD 4870, Australia. Email: xi.zhou1@my.jcu.edu.au, tao.huang1@jcu.edu.au.}
\thanks{Q.-L.~Han is with the School of Science, Computing and Engineering Technologies, Swinburne University of Technology, Melbourne, VIC 3122, Australia. Email: qhan@swin.edu.au.}
}

\begin{document}

\markboth{IEEE Robotics and Automation Letters}%
{\MakeLowercase{\textit{et al.}}: Demo of IEEEtran.cls for IEEE Journals}

\maketitle

\begin{abstract}

Cooperative perception (CP) is essential to enhance the efficiency and safety of future transportation systems, requiring extensive data sharing among vehicles on the road, which raises significant privacy concerns. 
Federated learning offers a promising solution by enabling data privacy-preserving collaborative enhancements in perception, decision-making, and planning among connected and autonomous vehicles (CAVs).
However, federated learning is impeded by significant challenges arising from data heterogeneity across diverse clients, potentially diminishing model accuracy and prolonging convergence periods. 
This study introduces a specialized federated learning framework for CP, termed the federated dynamic weighted aggregation (FedDWA) algorithm, facilitated by dynamic adjusting loss (DALoss) function. 
This framework employs dynamic client weighting to direct model convergence and integrates a novel loss function that utilizes Kullback-Leibler divergence (KLD) to counteract the detrimental effects of non-independently and identically distributed (Non-IID) and unbalanced data. 
Utilizing the BEV transformer as the primary model, our rigorous testing on FedBEVT dataset which is expanded on OpenV2V dataset, demonstrates significant improvements in the average intersection over union (IoU). 
These results highlight the substantial potential of our federated learning framework to address data heterogeneity challenges in CP, thereby enhancing the accuracy of perception models and facilitating more robust and efficient collaborative learning solutions in the transportation sector.
\end{abstract}


\section{Introduction}

Environmental perception is crucial for advanced driver assistance systems (ADAS) and autonomous driving, encompassing tasks like instance segmentation \cite{xiong2022contrastive,liu2022deep,10088, Radar_Ins}, traffic scene understanding \cite{10086}, 2D object detection \cite{10087}, 3D object detection \cite{GA_3DOD, SMURF, LXL, yang2022ralibev}, and multi-object tracking \cite{GNN_PMB, liu2023framework, ShaSTA, ding2024lidar}. 
These technologies also play a critical role in enhancing cooperative intelligent transportation systems (C-ITS), significantly improving transportation safety and efficiency \cite{huang2023v2x, hu2023collaboration, li2023among}. 
The rise of C-ITS has drawn considerable interest from both academia and industry, highlighting a collaborative framework that integrates various ITS components—personal, vehicle, roadside, and central systems—to surpass traditional ITS models in service quality. 
Connected automated vehicles (CAVs) exchange information through vehicle-to-everything (V2X) networks, including vehicle-to-vehicle (V2V), vehicle-to-infrastructure (V2I), vehicle-to-network (V2N), and infrastructure-to-network (I2N).


A major challenge within C-ITS is the security and privacy of shared data. 
Federated learning in general offers a robust solution by enabling decentralized model training without compromising data privacy, which has already been demonstrated in many applications like wireless communications \cite{Fed-WC}, Industry 4.0. \cite{Fed-Ind} and robotics \cite{Fed-HANet}, etc.
In C-ITS, this approach also allows CAVs to collaboratively participate in environment perception, traffic flow prediction, and decision-making, to improve deep neural network (DNN) models while keeping the privacy protected by avoiding exchange sensor data or latent features between CAVs \cite{Survey_FL_CP, Survey_FL_CP_TIV}.
Additionally, V2X communication supports seamless data exchange between vehicles, roadside units (RSUs), and cloud systems, enhancing the effectiveness of federated learning in C-ITS \cite{cress2021intelligent, huang2024RFLST, huang2020}. 
Supported by RSUs, digital twins can produce accurate labels for learning algorithms, thereby increasing the precision and performance of the systems.


Despite its numerous advantages, the federated learning paradigm encounters a significant challenge in data heterogeneity when implemented across diverse clients. 
This issue is particularly pronounced in C-ITS perception which is also known as cooperative perception (CP) due to the varied sensor configurations among participating vehicles and infrastructure.
Such variations include discrepancies in camera counts, sensor placements, and non-independently and identically distributed (Non-IID) data among traffic agents, leading to marked differences in data characteristics \cite{philion2020lift}. 

\textcolor{black}{In a CP situation, dealing with data heterogeneity and distribution shifts is crucial. These challenges can significantly impact the performance and reliability of the system. For example, some clients may have the same type of sensors on one hand. However, their configurations such as mounting positions, angles, or calibration, may vary. On the other hand, some clients may have different types of sensors, which may lead to different data characteristics. Specifically, the dataset supplied by FedBEVT \cite{song2023fedbevt} incorporates data from three distinct client types: cars, buses, and trucks. Each type of vehicle has sensors installed at varying positions and angles, and the calibration among these sensors also differs, resulting in data heterogeneity. Moreover, environmental variations further influence data distribution shifts. For example, changes in lighting conditions across multiple time frames within a single scene can significantly alter. Such phenomena could be observed clearly in the dataset supplied by FedBEVT \cite{song2023fedbevt}. 
}
These are significant challenges for traditional federated learning methods, as it may be challenging to effectively synchronize the contributions of local models towards the convergence of the global model.
This paper explores strategies to harness the benefits of external data while mitigating the impacts of data heterogeneity within a federated learning framework, with the aim of significantly improving local models' performance.


Among the critical tasks essential for CP, bird's-eye-view (BEV) perception is emphasized due to its key role in achieving comprehensive environmental awareness.
To address the challenge of data heterogeneity, we introduce a novel federated learning framework specifically designed for BEV perception.
Specifically, we proposed an advanced aggregation methodology within our federated learning framework, to confront with varied data distributions among clients and their disparate impacts on global model convergence.
This aggregation mechanism is designed to handle the heterogeneity intrinsic to diverse agents, thereby enhancing training efficiency and model efficacy.
Additionally, we introduced a novel loss function designed to mitigate unstable convergence issues, steering the model toward alignment with the global optimizer.



To validate our approach with fair comparison, we conducted extensive experiments and follow the same BEV semantic segmentation network model proposed in \cite{song2023fedbevt}, which consists of five main components: an encoder for image feature extraction; positional embedding to capture camera geometry; a cross-attention module that transitions front views to BEV; convolution layers within transformers for feature refinement; and a decoder to convert BEV representations into actionable predictions for BEV semantic segmentation.
The results demonstrate the superiority of our framework, showing improved test accuracy and reduced communication overhead across most clients, thus confirming the effectiveness of the proposed federated learning framework in addressing the complex challenges.


Our contributions are summarized as follows:
\begin{itemize}
\item A federated learning framework tailored for CP task with a novel federated dynamic weighted aggregation (FedDWA) algorithm is introduced.
This framework addresses data heterogeneity across clients by employing a dynamic constant mechanism, which adjusts the balance between global and local model, narrowing the performance gap and fostering a cohesive learning environment.
\item To enhance training efficacy further, a dynamic adjusting loss (DALoss) function is proposed for CP task under federated learning framework. 
This function is tailored to the varying data distributions among clients and the central server, fine-tuning the model's convergence direction based on real-time data distribution insights, thus ensuring improved convergence and significantly boosting the accuracy of BEV perception outcomes.
\end{itemize}

\section{Related Work}\label{related_work}

\subsection{Cooperative Perception in Autonomous Driving}

It is essential that autonomous driving systems have the ability to perceive the environment accurately. 
This perception directly affects the efficiency of decision-making and overall safety.
Traditional single vehicle perception systems, \textcolor{black}{although have recently been equipped by the BEV perception paradigm which extracts BEV space feature from input sequences in perspective view or 3D space, facilitate performance of perception tasks such as 3D bounding box detection and BEV segmentation \cite{BEVFormer, BEVDet, BEVFusion},} yet still often struggle with occlusions and distant objects since being constrained by limited sensing capabilities. 
CP, enabled by V2X communications, marks a significant improvement, allowing CAVs to enhance perception through extensive wireless information sharing, encompassing V2V, V2I, and V2N communications. 
OpenCDA, introduced by \cite{xu2021opencda}, provides a versatile framework for developing and testing cooperative driving automation (CDA) systems within a simulated environment. 
Complementarily, the work \cite{xu2022opv2v} presented a large-scale simulated dataset for V2V perception, enriched with a diverse collection of scenes and annotated data. 
This dataset proves critical for advancing CP systems, as supported by \cite{huang2023v2x}, which underscores the expanded sensing capabilities facilitated by V2X. 
Further developments include the CoBEVT framework introduced in \cite{xu2022cobevt}, a sparse vision transformer tailored for autonomous driving that leverages multi-agent multi-camera sensors, enabling collaborative perception across multiple vehicles. 
The work \cite{yang2021machine} discusses the integral role of advanced algorithms in optimizing CP systems. 
The works \cite{bai2022infrastructure} and \cite{han2023collaborative} explore the potential of integrating infrastructure-based sensors to bolster CP. 
There have been some notable advancements in perception frameworks recently. 
One of these is the introduction of V2X-ViT \cite{xu2022v2x}. 
V2X-ViT is a novel vision Transformer designed to effectively combine information from on-road agents. 
Another important contribution is the work by Xu et al. \cite{xu2023v2v4real}. 
They have created the first extensive real-world multimodal dataset for V2V perception. 
This dataset significantly enhances data availability for CP research.
Lastly, Xiang et al. have introduced HM-ViT in their paper \cite{xiang2023hm}. 
HM-ViT is an innovative multi-agent hetero-modal cooperative perception framework. 
It can accurately predict 3D objects in dynamic V2V scenarios.
Collectively, these advancements represent a major leap towards more comprehensive and precise environmental perception in autonomous driving, addressing critical challenges and enhancing driving safety.

\vspace{-5pt}
\subsection{Federated Learning}

Federated learning, initially proposed as FedAvg by \cite{mcmahan2017communication}, marks a significant shift in machine learning by enabling distributed model training across multiple systems without accessing users' raw data, thus enhancing data privacy. 
A primary challenge in federated learning is the management of data heterogeneity and distribution shifts, which complicate efficient model optimization. 
The work \cite{li2020federated} introduces a method to mitigate client heterogeneity by incorporating proximal terms, ensuring that local model updates align with the global model. 
Additionally, the work \cite{lin2020ensemble} utilizes knowledge distillation techniques to aggregate locally computed logits, allowing for the synthesis of cohesive global models without requiring uniform local model architectures. 
Further addressing this heterogeneity, the works \cite{wang2021novel, li2020federated} propose algorithms that adjust the training objective to tackle these challenges effectively. 
The work \cite{karimireddy2020scaffold} introduced SCAFFOLD, which employs control variables for variance reduction to correct ‘client-drift’ in local updates. 
Moreover, the work \cite{qu2022generalized} presents FedSAM, an algorithm that integrates sharpness aware minimization (SAM) as a local optimizer with a momentum-based federated learning algorithm to enhance synergy between local and global models.

\vspace{-5pt}
\subsection{Federated Learning with Cooperative Perception}

In CP involving CAVs and RSUs, the exchange of sensor data raises significant privacy concerns. This data, sourced from vehicles and smart city infrastructure, can reveal sensitive information about individuals, such as location and personal habits. 
To address this, the work \cite{jallepalli2021federated} applied federated learning to object detection tasks in AVs, capitalizing on the method's ability to train models without compromising data privacy. 
However, the broader application of federated learning in CP was initially underexplored.
Recent studies such as \cite{Survey_FL_CP}, \cite{Survey_FL_CP_TIV} and \cite{CP_with_Reliable_FL} discuss the application of federated learning in CP settings. 
The work \cite{song2022federated} introduces an innovative federated learning framework, H2-Fed, designed to handle Hierarchical Heterogeneity, significantly improving conventional pre-trained deep learning models for CP. 
Furthermore, the work \cite{song2023v2x} proposes a new client selection pipeline using V2X communications, optimizing client selection based on predicted communication latency in real CP scenarios. 
These developments underscore the potential for federated learning for collaborative model training in privacy-sensitive environments.
Despite these advances, implementing CP through federated learning presents challenges, primarily due to heterogeneity in data distribution, computational resources, and network connectivity among devices. 
This variability can significantly affect learning efficiency and effectiveness. 
To address these challenges, the work \cite{tang2023fair} introduces a fair and efficient algorithm to manage the imbalanced distribution of data and fluctuating channel conditions. 
Moreover, the work \cite{song2023fedbevt} proposed FedBEVT, a federated transformer learning approach that focuses on BEV perception to tackle common data heterogeneity issues in CP.

\section{Methodology}
\label{med}

\subsection{Problem Formulation}

In this paper, we aim to establish a federated learning framework, $FL$, consisting of $M$ clients that each equipped with a unique label $m\in\{1,2,...,M\}$, for the BEV semantic segmentation task under V2X CP setting.
The $m$-th client contains a set of data samples denoted by $\mathbf{N}_m=\{(\mathbf{N}_{m,i},\mathbf{Y}_{m,i})\}_{i\in\{1,2,...,|\mathbf{N}_m|\}}$, where $|\cdot|$ denotes the number of elements in the set, $\mathbf{N}_{m,i}$ is the set of images in the $i$-th data sample captured from a multi-view camera system, and $\mathbf{Y}_{m,i}$ is the set of corresponding BEV semantic masks. 
Within this framework, each client $m$ trains a local model $G_m$ independently using the calculated gradient, by minimizing a primary optimization objective loss function $L_{m,i}$ for each data sample in their dataset, and then transmits the model updates to a central server. Thus this loss function is crucial for assessing and minimizing the error throughout the distributed network of clients $m$. 
The server integrates these updates to improve the global model.
This cycle of local training and central updating repeats across several communication rounds to refine the global model optimally.



To demonstrate the effectiveness of our proposed federated learning framework, we follow FedBEVT \cite{song2023fedbevt} to employ the same transformer-based BEV segmentation network as network model $G_m$ and $G_g$, on each client and server respectively.
Such network model comprises an image feature encoder utilizing a CNN, a BEV transformer with a multi-layer attention structure (incorporating both sparse cross-view and self-attention mechanisms), and a BEV decoder. 
This configuration effectively encodes images from multiple cameras and transforms them into BEV features. 
These features are decoded to produce semantic segmentation outputs on the BEV perspective.


Our framework introduces a novel aggregation algorithm and a customized loss function designed to tackle two primary challenges for CP setting: 
(i) the heterogeneity of data and distribution shifts across different clients, and 
(ii) the varying contributions of different clients to the convergence of the global model. 
These enhancements are vital to ensure robust and precise model performance in autonomous driving environments.

\subsection{Federated Dynamic Weighted Aggregation}

To address the specific challenges of applying federated learning in CP, particularly client drift due to diverse sensor types and Non-IID data distributions, we introduce the federated dynamic weighted aggregation (FedDWA) algorithm for our federated learning framework $FL$. 
This algorithm updates the client model described by:
\begin{equation}
\label{eq2}
    G_m=G_m^{-}-\frac{1}{|\mathbf{N}_m|}U_m,
\end{equation}
\begin{equation}
\label{eq25}
    U_m = \eta_{m}(\sum_{i}^{|\mathbf{N}_m|} \partial_i(G_m^{-}) + c_g^- - c_m^-),
\end{equation}
\begin{algorithm}
    \small
    \SetAlgoLined 
    \caption{FedDWA}\label{algorithm_1}
    \SetKwInOut{Input}{Input}
    \SetKwInOut{Output}{Output}
    \Input{}
    Number of communication round $R$ and data samples $\mathbf{N}_m$ \leavevmode \\
    Server input: global model $G_{g}^{-}$ and global control variable $c_{g}^{-}$ in previous round, global learning rate $\eta_{g}$\leavevmode \\
    $m$-th client input: local control variable $c_{m}^{-}$ in previous round and local step-size $\eta_{m}$\leavevmode \\
    \SetKwFunction{MyFuns} {MyFuns}
    \SetKwProg{Fn}{Function}{:}{}
    \For{each round $r = 1, \dots, R$ }{
    \For{}{
    
    send $G_{g}^{-}$, $c_{g}^{-}$ to all clients $m \in  \{1,2,...,M\}$ \leavevmode \\
    \For{ client $m \in \{1,2,...,M\}$ parallel}{
    initialize local model on $m$-th client $G_{m} \gets G_{g}^{-}$ \leavevmode \\
    initialize $U_m = 0$\leavevmode \\
    initialize $W_m = 0$\leavevmode \\
    initialize $O_m = 0$\leavevmode \\

    \For{each data sample $\mathbf{N_{m,i}}$} {
    
    \tcp{\emph{Local model training}}\label{cmt}
    calculate mini-batch gradient  $\partial_i(G_m)$ \leavevmode \\
    $U_m$ $+=$ $\partial_i(G_m^{-})$\leavevmode \\
    $W_m$ $+=$ $\partial_i(G_m)$ \leavevmode \\
    $O_m$ $+=$ $\mathcal{D}_\mathrm{KL}\left(P_g^-(\mathbf{N}_{m,i}),P_m(\mathbf{N}_{m,i})\right)$

    }
    $U_m$ $=$ $\eta_m(U_m + c_g^- - c_m^-)$\leavevmode \\
    $G_m=G_m^{-} - \frac{1}{|\mathbf{N}_m|} U_m$ \leavevmode \\
    \tcp{\emph{Local control variable update}}\label{cmt}
    $c_{m} = \frac{1}{|\mathbf{N}_m|}  W_m$ \leavevmode \\
    $T_{m} = \frac{1}{|\mathbf{N}_m|}c_m\cdot O_m$ \leavevmode \\
    send $G_m$, $T_{m}$ to global server \leavevmode \\

    }
    
    }
    \tcp{\emph{Global control variable update.}}\label{cmt} \leavevmode \\
    $c_g=c_g^- + \frac{1}{M}\sum_{m=1}^M T_m$ \leavevmode \\
    \tcp{\emph{Global model update.}}\label{cmt} \leavevmode \\
    $G_g=G_g^{-}+\frac{\eta_g}{M}\sum_{m=1}^M (G_g^{-} - G_m)$  \leavevmode \\
    
    } 
\end{algorithm}
where $G_m$ represents (the weights of) the local model of the $m$-th client, $\eta_{m}$ is the local step-size, and $\partial_i(G_m^{-})$ denotes the gradient of the $i$-th data sample for the local model in the previous communication round. 

Each client uses a local control variable $c_m$ to account for variances in model updates, aligning local objectives with the global model to mitigate client-specific drifts, and $c_m^-$ denotes the local control variable in the previous communication round. Both $c_m$ and $c_g$ are initialized as 0.

Inspired by \cite{karimireddy2020scaffold}, we maintain a global control variable $c_{g}$ on the server, aggregating information from all clients to guide the overall direction of model updates. 
Diverging from \cite{karimireddy2020scaffold}, which uses the average of local variables to update $c_{g}$, we utilize Kullback-Leibler divergence (KLD) \cite{van2014renyi} to dynamically weigh the contributions of different clients like:

\vspace{5pt}
\begin{equation}
\label{eq11}
    c_g=c_g^- + \frac{1}{M}\sum_{m=1}^M T_m,
\end{equation}
where $T_m$ is the intermediate variable, as shown in:
\begin{equation}
    T_m = \frac{1}{|\mathbf{N}_m|}c_m\cdot O_m,
\end{equation}
and
\begin{equation}
     O_m = \sum_{i}^{|\mathbf{N}_{m}|} \mathcal{D}_\mathrm{KL}\left(P_g^-(\mathbf{N}_{m,i}),P_m(\mathbf{N}_{m,i})\right).
\end{equation}

The KLD is calculated between the predicted data distribution of the global model from the previous communication round, $P_{g}^{-}(\mathbf{N}_{m,i})$, and the distribution of $m$-th local client, $P_{m}(\mathbf{N}_{m,i})$. Where $P_{g}^{-}(\mathbf{N}_{m,i})$ is approximated by using the segmentation output of global model from previous communication round $G^{-}_{g}$, and $P_{m}(\mathbf{N}_{m,i})$ is approximated by using the segmentation output of each local model $G_{m}$.
\textcolor{black}{To transform the segmentation output into a distribution,  a log softmax function is applied to convert the segmentation output into log probabilities, e.g.: 
}
\textcolor{black}{
\begin{equation}
\begin{split}
    P_{m}(\mathbf{N}_{m,i}) = \begin{bmatrix}
  log(\sigma (V_{1,1}))& \dots & log(\sigma (V_{1,k}))\\
  \dots &  \dots & \dots \\
   log(\sigma (V_{j,1}))&  \dots & log(\sigma (V_{j,k}))
\end{bmatrix},
\end{split} 
\label{eq126}
\end{equation}
}
\textcolor{black}{where $V_{j,k}$ denotes the $(j,k)$-th pixel values in the segmentation output matrix $G_{m}(\mathbf{N}_{m,i})$ and $\sigma$ represents the softmax function. The same calculation is applied for $P_{g}^{-}(\mathbf{N}_{m,i})$}. 

%
%
Each client uses gradients to update the local control variable $c_{m}$:
\begin{equation}
    \label{eq666}
    c_{m} = \frac{1}{|\mathbf{N}|_m}  W_m,
\end{equation}
where
\begin{equation}
    W_m = \sum_{i}^{|\mathbf{N}_{m}|} \partial_i(G_m).
\end{equation}


The global model is updated as outlined below:
\begin{equation}
\label{eq33}
G_g=G_g^{-}+\frac{\eta_g}{M}\sum_{m=1}^M (G_g^{-} - G_m),
\end{equation}
where $\eta_{g}$, the learning rate, plays a crucial role in balancing the updates, and $G_{g}$ represents the global model. 
This equation captures the aggregation of differences between the global and local models, promoting the convergence of the global model. 
To provide an comprehensive understanding of FedDWA, the pseudo code is illustrated as \textbf{Algorithm \ref{algorithm_1}}.

\subsection{Dynamic Adjusting Loss Function}


As discussed before, it is critical to define a suitable loss function $L_{m,i}$, to calculate the gradient of $i$-th data sample for each client's local model, $\partial_i(G_m)$. The cross-entropy loss function is widely used within federated learning frameworks to train individual clients. 
This function effectively gauges the performance of classification models that output probabilities between 0 and 1. 
Nevertheless, the inherent data heterogeneity across clients, particularly with non-IID data, complicates the convergence to a global optimum.



To address this issue, we introduce an enhanced local training loss function termed dynamic adjusting loss (DALoss):
\begin{equation}
\label{eq444}
L_{m,i}=\mathcal{S}\left(G_{m}(\mathbf{N}_{m,i}),\mathbf{Y}_{m,i}\right)+Q_{m,i},
\end{equation}
where $\mathcal{S}\left(G_{m}(\mathbf{N}_{m,i}),\mathbf{Y}_{m,i}\right)$ denotes the standard cross-entropy loss for the $i$-th data sample of client model $G_{m}$ against the true labels $\mathbf{Y}_{m,i}$, and $Q_{m,i}$ is the dynamic control term defined by:
\begin{equation}
\label{eq3333}
    Q_{m,i} = C\cdot \mathcal{D}_\mathrm{KL}\left(P_g^-(\mathbf{N}_{m,i}),P_m(\mathbf{N}_{m,i})\right) \cdot||G_g^{-} - G_m||_2^2,
\end{equation}
where $C\cdot \mathcal{D}_\mathrm{KL}\left(P_g^-(\mathbf{N}_{m,i}),P_m(\mathbf{N}_{m,i})\right)$ scales the quadratic penalty based on the KLD, measuring the discrepancy between the probabilistic distributions of the global and local models from the previous communication round. 
$C$ is a regularization constant, and  $||\cdot||_2$ denotes the $L^2$ norm. This divergence quantifies the differences, ensuring that local updates contribute significantly toward global model convergence. 

This modification incorporates a dynamic adjustable regularization mechanism to better align each client’s model updates with the global model. 
Such dynamic adjustment of the regularization parameter allows our approach to adapt to the evolving nature of the training process and the unique characteristics of each client.

\textcolor{black}{By integrating KLD into the loss function, the learning process is encouraged to minimize deviations between data distribution in different clients, effectively aligning the local models closer to the global model. Specifically, the KLD is employed as a weighting factor in the L2-norm calculation between the global and local models, as shown in Eq. (\ref{eq3333}). A large KLD indicates a significant discrepancy in the distributions, resulting in an increased weight. This increment in weight amplifies the loss, consequently accelerating the convergence of the local model towards the global model. By minimizing this divergence between local updates and the global model, the learning process adaptively adjusts to the unique data distributions present in each client. Therefore, our approach ensures that local updates contribute positively towards a more generalized global model, mitigating the impact of data distribution shifts and data heterogeneity.}

\textcolor{black}{
While the design of the loss function allows for larger updates when there are significant discrepancies between the local and global models, it is crucial to manage the potential instability might be introduced by these aggressive updates. To avoid such instability, we use a regularization constant $C$, which can penalize large changes in model weights and provide a counterbalance to aggressive updates. We also use adaptive learning rates which can adjust the step size based on the training dynamics that we reduce the learning rate after several aggregation steps thus tempering the update magnitude.
}

\vspace{-5pt}
\section{Experiments and Analysis}
\label{exp}


The experiments were conducted on a computer equipped with a single NVIDIA GeForce RTX 4090 GPU and a 12th Generation Intel Core i9-12900K CPU. 
The software environment consisted of Pytorch 2.0.1 and CUDA 12.3. \textcolor{black}{Our experiments are implemented on BEV semantic segmentation, a crucial task in autonomous driving technologies. BEV semantic segmentation facilitates a comprehensive top-down environmental perspective, which is essential for navigation, obstacle avoidance, and path planning in autonomous vehicles.}

\subsection{Experiment Setup}

\textcolor{black}{We do the experiment on the dataset provided by FedBeVT \cite{song2023fedbevt}, which extends the original OpenV2V \cite{xu2022opv2v} dataset only contained car objects by adding two types of vehicles: truck and bus with different camera sensor installation positions. This enhancement broadens the diversity and volume of the scenarios represented, providing a robust foundation for testing our model.}
\begin{table*}[ht]
\caption{Comparison between Our Method and Other State-of-the-art Methods.} 
\label{weightmatrix}
\renewcommand\tabcolsep{3.6pt}
\begin{center}
\begin{threeparttable}

\begin{tabular}{ |c|ccc|ccc|ccc|ccc|}
  \hline
  \multirow{3}{*}{Method} & \multicolumn{3}{c|}{Client 1: Bus}  & \multicolumn{3}{c|}{Client 2: Truck} & \multicolumn{3}{c|}{Client 3: Car A} & \multicolumn{3}{c|}{Client 4: Car B}\\
  \multicolumn{1}{|c|}{}& \multicolumn{3}{c|}{N1 = 1388}  & \multicolumn{3}{c|}{N2 = 1488} & \multicolumn{3}{c|}{N3 = 2140} & \multicolumn{3}{c|}{N4 = 1384}\\
  \multicolumn{1}{|c|}{}& Loss$\downarrow$ & IOU$\uparrow$ & Com.$\downarrow$   & Loss$\downarrow$ & IOU$\uparrow$ & Com.$\downarrow$  & Loss$\downarrow$ & IOU$\uparrow$ & Com.$\downarrow$ & Loss$\downarrow$ & IOU$\uparrow$ & Com.$\downarrow$ 
\\
\hline
Local Training & 0.0716 & 5.42$\%$ &-& 0.1740 & 4.16$\%$ & - & 0.0288 & 13.15$\%$ & - & 0.0298 & 9.41$\%$ & - \\
FedAvg \cite{mcmahan2017communication} & 0.2202 & 5.89$\%$ & \textbf{105} & 0.0483 & 5.89$\%$ & 180 & 0.0518 & 14.09$\%$ & \textbf{175} & 0.0899 & 14.61$\%$ & 175 \\
FedRep \cite{collins2021exploiting} & 0.0170 & 7.45$\%$ & 320 & 0.1319 & 7.03$\%$ & 105 & 0.0352 & 18.94$\%$ & 345 & 0.0394 & 15.41$\%$ & 330\\
FedTP \cite{li2023fedtp} & 0.0765 & 6.73$\%$ & 175 & 0.1015 & 5.42$\%$&140 & 0.0183 & 17.29$\%$ & 335 &0.0389 & 15.56$\%$ & 300\\
FedBEVT \cite{song2023fedbevt} & \textbf{0.0061} & 10.32$\%$ & 340 & 0.1294 & 7.33$\%$ & \textbf{65} & 0.0162 & 18.40$\%$ & 360  &0.0268 & 16.16$\%$ & 370\\
\hline
FedDWA with DALoss (Ours) & 0.0062 &\textbf{10.72$\%$} &235 & \textbf{0.0094} &\textbf{15.91$\%$} &105 & \textbf{0.0157} &\textbf{21.30$\%$} &484 & \textbf{0.0198} &\textbf{19.35$\%$} &\textbf{100}\\
\hline
\end{tabular}
\begin{tablenotes}
        \footnotesize
        \item[$\uparrow$] The upper arrow denotes that better performance is registered with higher value.
        \item[$\downarrow$] The lower arrow denotes that better performance is registered with lower value.
\end{tablenotes}
\end{threeparttable}
\end{center}
\vspace{-1.0em}
\end{table*}

Following the protocol established in \cite{song2023fedbevt}, our study orchestrates a collaborative framework among four distinct industrial clients to assess the efficacy of our proposed object detection model under various operational conditions  Specifically, the dataset comprises:
\begin{itemize}
    \item \textbf{Client 1 Bus}: 1,398 frames in the training set and 413 frames in the testing set.
    \item \textbf{Client 2 Truck}: 1,459 frames in the training set and 356 frames in the testing set.
    \item \textbf{Client 3 Car A}: 11,636 frames in the training set and 3,546 frames in the testing set.
    \item \textbf{Client 4 Car B}: 7,142 frames in the training set and 10,191 frames in the testing set.
\end{itemize}


For a fair comparison between our federated learning strategy and state-of-the-art approach FedBEVT \cite{song2023fedbevt}, our model architecture adheres to the configuration described in \cite{song2023fedbevt}. 
This involves processing camera images observed by clients. 
Similar to \cite{song2023fedbevt}, we begin by sending these images through a 3-layer ResNet34 encoder, which is proficient at extracting and encoding image features across various spatial dimensions. 
This encoding process yields feature tensors at differentiated spatial resolutions: $64 \times 64 \times 128$, $32 \times 32 \times 256$, and $16 \times 16 \times 512$. 
\begin{figure}[htbp]

	\centering
	\begin{subfigure}{0.49\linewidth}
		\centering
		\includegraphics[width=0.9\linewidth]{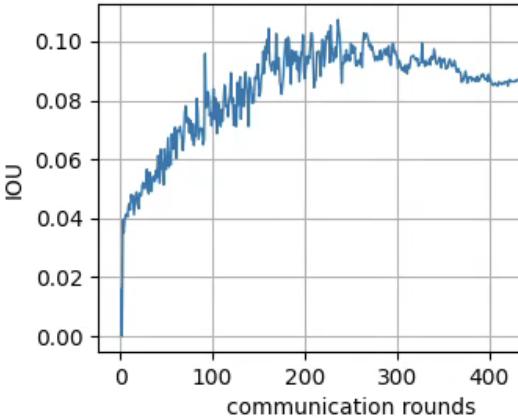}
		\caption{Bus client}
	\end{subfigure}
 \hfill
	\begin{subfigure}{0.49\linewidth}
		\centering
		\includegraphics[width=0.9\linewidth]{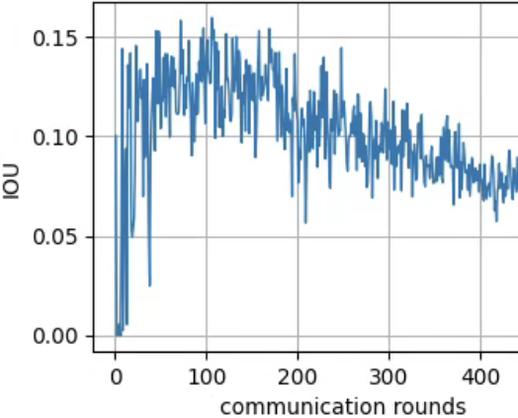}
		\caption{Truck client}
		\label{chutian2}
	\end{subfigure}
	\qquad
 
	\centering
	\begin{subfigure}{0.49\linewidth}
		\centering
		\includegraphics[width=0.9\linewidth]{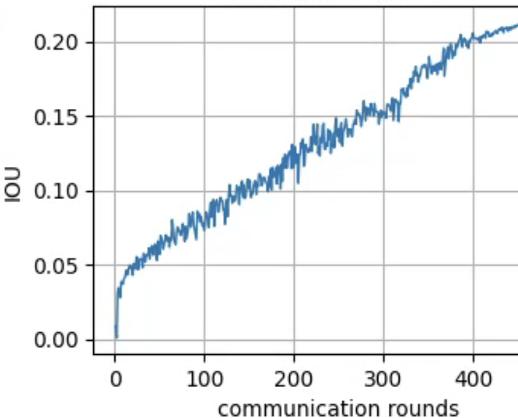}
		\caption{Car client A}
		\label{chutian3}
	\end{subfigure}
 \hfill
	\begin{subfigure}{0.49\linewidth}
		\centering
		\includegraphics[width=0.9\linewidth]{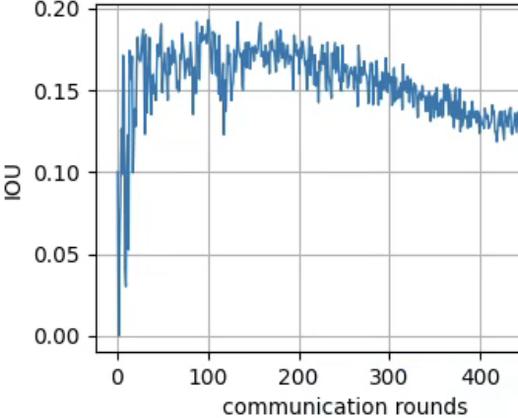}
		\caption{Car client B}
		\label{chutian4}
	\end{subfigure}
 \caption{This figure illustrates the results for four distinct clients. The x-axis represents the number of communication rounds, while the y-axis indicates the IoU values. Subfigure (a) displays the performance for the bus client, subfigure (b) shows the results for the truck client, and subfigures (c) and (d) present the outcomes for car client A and car client B, respectively.}
 \label{figg}
\end{figure}
Subsequently, we employ a focused attention cross (FAX) transformer operation, which is attention-based and skillfully manages interactions between BEV embeddings and the encoded image features, conceptualized as query, key, and value within this framework. 
Finally, a 3-layer bilinear module is used to transform these enriched BEV features.
In our model configuration, the hyper-parameter $C$ defined in Equation (\ref{eq3333}) is set to 0.1. For other hyper-parameters, we adopt the configurations detailed in \cite{song2023fedbevt}. We selected these settings due to their proven effectiveness in similar contexts, ensuring that our model benefits from established best practices while maintaining consistency with state-of-the-art for fair comparison.

\vspace{0pt}
\subsection{Performance Analysis}


In our comparative study, we evaluate the performance of our proposed algorithm, FedDWA, against several established methods, including FedAvg \cite{mcmahan2017communication}, FedRep \cite{collins2021exploiting}, FedTP \cite{li2023fedtp}, and FedBEVT \cite{song2023fedbevt}. 
The evaluation focuses on the average intersection over union (IoU) achieved by models across three use cases (UCs) and the number of communication rounds required to attain optimal IoU values.

\textcolor{black}{The pixel-wise IoU metric is a common evaluation metric for BEV semantic segmentation. It quantifies the overlap between the predicted segmentation and the ground truth, providing a measure of accuracy at the pixel level. The IoU is calculated as follows:}

\vspace{0pt}
\textcolor{black}{
\begin{equation}
    IoU =  \frac{TP}{TP + FP + FN},
\end{equation}
}

\textcolor{black}{where TP is the number of true positive pixels, FP is the number of false positive pixels, and FN is the number of false negative pixels.
}


The experimental analysis presented in Table \ref{weightmatrix} provides a comprehensive comparison of our federated learning framework against these prominent approaches. 

Performance metrics include training loss, IoU, and the number of communication rounds (Com.) required to achieve peak IoU, offering a detailed assessment of the efficacy and efficiency of our approach within a federated learning context.


The key findings for each client scenario are as follows:

\begin{itemize}
\item \textbf{Client 1 Bus}: Our method shows a significant improvement, achieving a 10.72$\%$ IoU after 235 communication rounds, demonstrating effective data heterogeneity management and communication efficiency.

\item \textbf{Client 2 Truck}: Our approach excels, reaching a 15.91$\%$ IoU, marking a notable improvement over local training and indicating robust adaptability and superior management of client-specific data characteristics.
\item \textbf{Client 3 Car A}: Achieving the highest IoU of 21.30$\%$ among all methods, our approach excels in complex vehicular environments, underscoring its potential to optimize the federated learning framework across diverse datasets.
\item \textbf{Client 4 Car B}: This client achieves a 19.35$\%$ IoU in just 100 communication rounds, highlighting both the efficacy and efficiency of our approach and establishing it as a leading method in federated learning for BEV perception.
\end{itemize}


Fig. \ref{figg} illustrates the relationship between the number of communication rounds and IoU during the testing phase for the four clients, providing a detailed view of test progression. 
Notably, the truck client and car client B reached optimal IoU with fewer communication rounds, potentially due to data distributions more closely aligned with the predictions of their local models. 
However, after reaching peak IoU, there is a noticeable decline in performance, suggesting possible overfitting as the models become excessively tailored to their local datasets, to the detriment of generalization.

\subsection{Ablation Experiment}

\begin{figure*}[htbp]
\centering
\includegraphics[width=\textwidth]{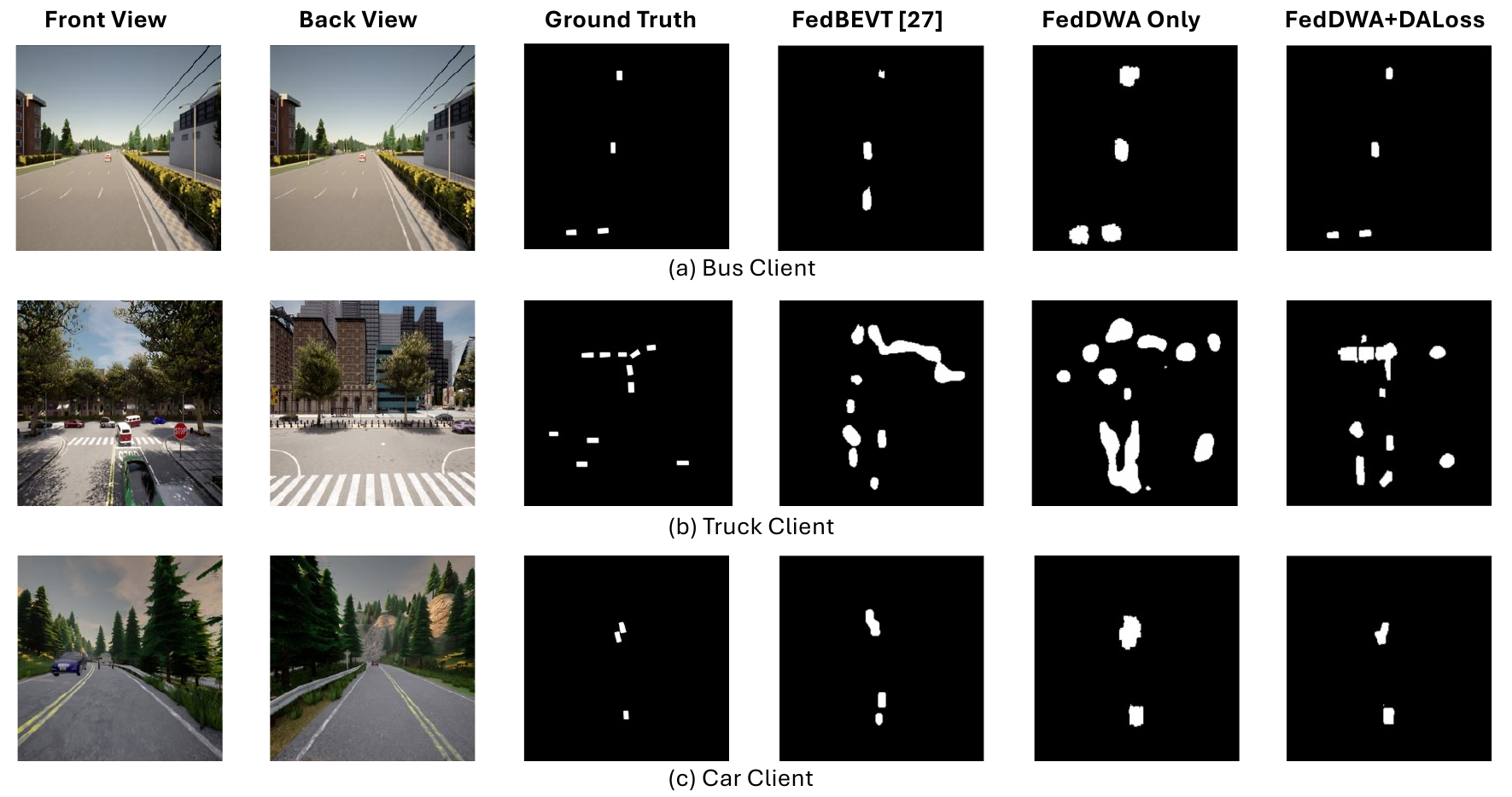} 
\caption{\textcolor{black}{This figure is the visualization of model output across different clients. The ego vehicle is on the center. Each panel in this figure presents comparative results for different vehicle clients: (a) Bus, (b) Truck, and (c) Car. For each subfigure, data is organized into six columns. The first column displays camera imagery from the frontal perspective, while the second column shows the rear perspective. The third column represents the ground truth. The fourth column illustrates model output from FedBEVT \cite{song2023fedbevt}. The fifth column shows the result from FedDWA without DALoss. The final column depicts result from the model trained with both FedDWA and DALoss, showcasing the effectiveness of the proposed method.}}
\label{vis}
\vspace{-1.0em}
\end{figure*}

\begin{table}
\caption{Ablation Experiment for Effectiveness of DALoss with IoU}
\centering
\begin{tabular}{|c|c|c|c|c|} 
\hline 
Method & Client 1 & Client 2 & Client 3 & Client 4 \\
\hline

FedBEVT \cite{song2023fedbevt}  & 10.32$\%$ & 7.33$\%$ & 18.40$\%$ & 16.16$\%$  \\

FedBEVT + DALoss & \textbf{10.49}$\%$ &\textbf{11.97}$\%$&\textbf{19.20}$\%$& \textbf{16.45}$\%$\\
\hline
FedDWA (ours)& 10.66$\%$& 14.88$\%$& 20.83$\%$& 18.28$\%$\\

FedDWA (ours) + DALoss  & \textbf{10.72}$\%$ &\textbf{15.91}$\%$&\textbf{21.30}$\%$& \textbf{19.35}$\%$\\

\hline

\end{tabular}
\vspace{-1.0em}
\label{table3}
\end{table}

The ablation study summarized in Table \ref{table3} elucidates the incremental impact of our DALoss and the intrinsic benefits of our proposed FedDWA framework. This study methodically analyzes the improvements offered by DALoss. The addition of DALoss consistently improves performance across all clients. When DALoss is added to FedBEVT \cite{song2023fedbevt}, there is a noticeable improvement in IoU percentages for Client 2 (from 7.33$\%$ to 11.97$\%$) and modest gains across other clients. This indicates that DALoss facilitates better generalization. Our proposed FedDWA framework, even without DALoss, shows better performance than FedBEVT. For example, FedDWA achieves an IoU of 20.83$\%$ for Client 3 and 18.28$\%$ for Client 4, which are significant improvements over FedBEVT's 18.40$\%$ and 16.16, respectively. This enhancement underscores the effectiveness of our weighting approach in distributing model updates across different clients in a way that respects their unique data distributions. The combination of FedDWA and DALoss shows the most substantial gains in performance. For instance, Client 2's IoU increases from 14.88$\%$ with only FedDWA to 15.91$\%$ with the addition of DALoss, and similar incremental benefits are observed for other clients. The highest performance is observed in Client 3, where the IoU reaches 21.3$\%$. \textcolor{black}{Fig \ref{vis} illustrates the visualization of an example network model output across different clients. It compares the results among the ground truth, the current state-of-the-art approach FedBEVT \cite{song2023fedbevt}, the FedDWA only, and both FedDWA and DALoss on bus client, truck client and car client. The comparison shows that the model incorporating both FedDWA and DALoss achieves better performance.}

\section{Conclusion}\label{conclusion}


In this paper, we explore the effectiveness of federated learning frameworks for optimizing transformer-based models in BEV perception within road traffic datasets. 
A primary focus of our investigation is the challenge of data heterogeneity, which significantly impedes the performance of federated learning methodologies. 
To address this issue, we introduce an innovative algorithm, FedDWA, together with DALoss, specifically designed to mitigate the adverse effects of data diversity. 
Our comparative analyses with existing federated learning frameworks demonstrate that our proposed solutions significantly enhance the efficacy of the federated learning framework. 
These improvements are primarily achieved through dynamic adjustments of the model convergence direction and the equitable integration of diverse local models into a unified global model. 
Our findings underscore the viability and potential of federated learning frameworks in BEV perception, suggesting the approach as a resilient, efficient, and inclusive alternative for model training.





\end{document}